\documentclass[journal]{IEEEtran}
\pdfoutput=1

\usepackage[pdftex]{graphicx}
\usepackage{url}
\usepackage[table]{xcolor}

\usepackage{caption}
\usepackage{subcaption}

\newcommand{\REPORTARXIV}{\url{http://www.codyco.eu/survey-simulation}}

\begin{document}

\title{Tools for dynamics simulation of robots: a survey based on user feedback}

\author{Serena Ivaldi$^{\dag,\ddag}$, 
        Vincent Padois$^{\dag,\ddag}$ 
        and~Francesco Nori$^{\S}$ 
        \thanks{E-mail: serena.ivaldi@isir.upmc.fr}
\thanks{
$^\dag$ Sorbonne Universit\'es, UPMC Paris 06, UMR 7222, Institut des Syst\`emes Intelligents et de Robotique (ISIR), F-75005, Paris, France. }
\thanks{
$^\ddag$  CNRS, UMR 7222, Institut des Syst\`emes Intelligents et de Robotique (ISIR), F-75005, Paris, France. 
}
\thanks{$^\S$ Robotics, Brain and Cognitive Sciences Dept., Italian Institute of Technology.}
}

\markboth{Tools for dynamics simulation of robots - extended report}%
{Ivaldi \MakeLowercase{\textit{et al.}}: Tools for dynamic simulation of robots}

\maketitle

\begin{abstract}
The number of tools for dynamics simulation has grown in the last years.
It is necessary for the robotics community to have elements to ponder which of the available tools is the best for their research.
As a complement to an objective and quantitative comparison, difficult to obtain since not all the tools are open-source, an element of evaluation is user feedback.
With this goal in mind, we created an online survey about the use of dynamical simulation in robotics.
This paper reports the analysis of the participants' answers and a descriptive information fiche for the most relevant tools.
We believe this report will be helpful for roboticists to choose the best simulation tool for their researches.
\end{abstract}

\IEEEpeerreviewmaketitle

\section{Introduction}

With the progress of powerful computers enabling fast computations, dynamics simulation in robotics is no longer expected to be an offline computational tool.
It is used to rapidly prototype controllers, evaluate robots design, simulate virtual sensors, provide reduced model for model predictive controllers, supply with an architecture for real robot control, and so on.

There is a growing number of tools for dynamics simulation, ranging from dynamic solver libraries to systems simulation software, provided through either open or closed source code solutions, each more or less tailored to their expected domains of application.

The spectrum of robotics applications being large and in expansion, it is necessary for the developer community to have a feedback about the users' needs, and for the researchers to be aware of the available tools and have the elements to ponder which of the available tools is the best for their research.

With this goal in mind, we created an online survey about the use of dynamical simulation in robotics.
\footnote{Online survey: \url{http://goo.gl/Tmyf5A}}
The survey was divided into four parts: general information about the user, user experience with dynamics simulation in general, user experience with one tool of his choice, technical questions and subjective evaluation about the selected tool. 
The survey was advertised on the main robotics mailing lists (e.g., euron-dist, robotics-worldwide) as well as in other mailing lists of correlated disciplines (e.g. comp-neuro), and kept open for approximately one month. 

This paper summarizes 
the analysis of the users' answers. We also report a descriptive fiche for the most relevant software tools, for the reader's interest. 
In the appendix, we also report free comments about the subjective user experience (major problems and desiderata).

\subsection{Why user feedback?}

Most middleware for robotics (ROS, YARP, OROCOS, Player, etc.) are already open-source, some also cross-platforms. This makes it possible to produce interesting performance comparisons that can help the roboticists to pick the best middleware for their needs~\cite{Einhorn2012}.
Similar ideas (open-source and cross-platform compatibility) should be used to compare dynamics models and simulators.
For example, an interesting evaluation and performance comparison of contact modeling algorithms was presented in~\cite{Drumwright2011,Drumwright2012}.

As a complement to quantitative comparisons, a useful element of evaluation (often un-mentioned and neglected) is user feedback. What do users really think of the software they use for simulation? Would they suggest it? What is their experience in their particular use case? 
We believe user feedback may be useful to avoid time-consuming tuning and inappropriate choices of software to researchers. It could point a researcher to a community that is actively using the tool and that is sharing the same concern: for example, it is likely that people simulating flying robots have different needs than those simulating wheeled robots or those controlling bipeds.
Furthermore, user feedback can provide useful suggestions to the developers community about the things that matter the most to users in simulation.

\subsection{Challenges in simulation}

Dynamics simulators for robotics have more strict requirements than the ones used for animating virtual characters, where time, computational burden and physical reality can be less constraining.
In entertainment (e.g. video-games), unfeasible forces may not be a problem since the law of physics can be violated.
In  bio/mechanical studies, simulators can be used offline to analyze or synthesize behaviors. 
Although the field of dynamics modeling and simulation has 
matured over the last decades~\cite{Featherstone2008,Jain2011,Todorov2014}, the growing need to control whole-body movements of complex structures, such as humanoids, poses  additional challenges to simulators for robotics:

1) numerical stability, which poses strong limitations on the use of simulations in real-time control settings~\cite{Drumwright2011,Drumwright2012};

2) the capability to be used as predictive engines in real-time control loops~\cite{Todorov2012}, which requires the ability to be extremely fast in computing the dynamics and the guarantee for the solvers to converge to physically feasible solutions upon a certain time~\cite{Todorov2011};

3) the simulation of rigid and soft bodies in contact with rigid and compliant environments~\cite{Brogliato2002,Jia2013}: the inaccurate computation of contact forces between bodies may result in unrealistic contacts or physically unfeasible contact forces (this issue has been particularly evident in the virtual phase of the Darpa Robotics Challenge - DRC);

4) the capability to model and simulate new types of actuation systems, such as variable impedance or soft actuators~\cite{Duriez2013}, and different types of contacts, for example with deformable materials, compliant and soft surfaces~\cite{Duriez2006}.

Finally, the robotics community urges for standardized software tools and particularly open source software. 
The benefit of open-source is not only in the community that can grow around the software, developing new tools, improving its quality and avoiding to ``re-invent the wheel'' at each time, but also in checking its efficiency and robustness on real platforms (which is expensive).

\begin{figure*}
\begin{center}
\begin{subfigure}[b]{0.28\textwidth}
\centering
\includegraphics[height=4cm]{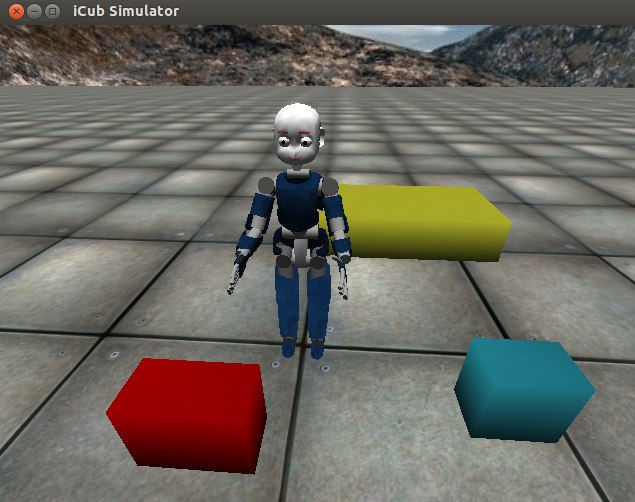}
\caption{iCubSim, based on ODE.}
 \label{fig:ode}
 \end{subfigure}
 \begin{subfigure}[b]{0.24\textwidth}
\centering
\includegraphics[height=4cm]{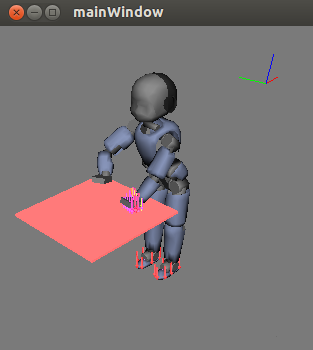} 
\caption{iCub in XDE.}
 \label{fig:xde}
 \end{subfigure}
\begin{subfigure}[b]{0.45\hsize}
\centering
\includegraphics[height=4cm]{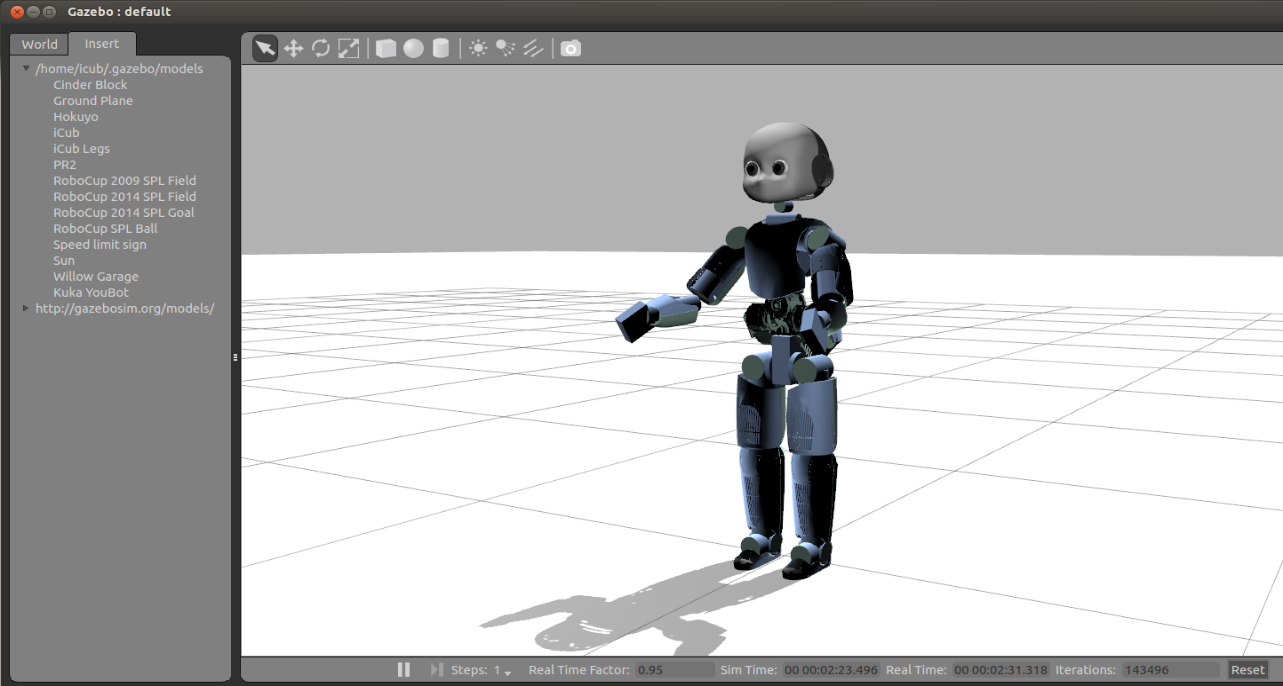}
\caption{iCub simulated in Gazebo.}
\end{subfigure}
\caption{Simulators of iCub. From left to right: iCubSim, based on ODE, XDE and Gazebo. (credits for Gazebo: Silvio Traversaro)}
\label{fig:simulators}
\end{center}
\end{figure*}

\subsection{The iCub case}

The iCub community recently faced the problem of choosing the correct tool for whole-body dynamics simulation. The existing simulator iCubSim~\cite{Tikhanoff2008}, is based on ODE and is mostly used as a tool for testing behaviors before trying them on the real robot. It is provided with an interface that emulates the low-level control of iCub, so the same code can be used to control simulated and real robot. 
However, the dynamics engine 
makes it inadequate for research about control of contacts and compliant surfaces.
At the moment two solutions are investigated: one based on XDE and the other based on Gazebo. The choice of these tools has been based on objective criteria (license, developing community, stability of the software simulation), previous experience and ``subjective feedback'' acquired orally discussing with colleagues, that provided partial and unstructured information. A more structured information about user feedback would have been helpful.
We believe this survey analysis could be a further element for choosing the best simulation tool in a research project.

\subsection{Comparing simulators}

It is certainly difficult to enumerate all the criteria that one can examine to choose a dynamics simulator, especially for a humanoid robot that is supposed to have physical interactions with rigid and compliant environments.

First, one can choose between physics engines (e.g. ODE, Bullet) and more complex softwares that include system simulation (e.g. Gazebo, V-Rep).

Second, facing the decision to adopt a simulator for a robot, a researcher should first decide between softwares that also include system simulation, and softwares which only simulate the dynamics of multi-body systems. This criterion allows us to consider under different perspectives two set of softwares: the first set, composed of software like Gazebo, OpenHRP, iCubSIM, which facilitate seamless simulation and control of the virtual characters and their corresponding physical system/robot; the second, like Humans, OpenSIM, Robotran, that are able to simulate the dynamics of complex systems but are not meant to provide seamless control of robotics platforms.

Another element of discrimination is the way the simulator represents rigid-body structures: on one hand we have software based on ODE and Bullet, such as Gazebo, iCubSim, MORSE, which represents joints as constraints between bodies; on the other we have softwares like XDE, OpenHRP, which make use of parameterized rigid-body dynamics representations, where joints are simply part of the robotics structure. These two classes determine not only the way forward/inverse dynamics are computed (and of course the second group also benefits from the straightforward computation of quantities useful in robotics, such as Jacobians, mass matrices etc.), but most importantly the way contact forces are computed. The first class considers contacts forces as bilateral/unilateral constraints, which are added to the list of constraints used to describe the joints; then the same solver is used to find the forces for the global system, including contacts and joints. In the second class, on the contrary, only constraints from the contacts are solved, which notably simplifies the problem. In generla, finding the correct contact forces can be burdensome. Current approaches to solve this problem are mostly based on the Linear Complementarity Problem (LCP)~\cite{Drumwright2012}, and in some cases there are mixed approaches combining LCP with optimization techniques, such as in MuJoCo~\cite{Todorov2012}. 

In short, there are several ``objective'' criteria that one can look at, on the basis essentially of what is advertised by the developers as a ``supported feature''. However, it is very difficult to find practical comparison of different simulators on test problems, for many reasons: first, an extensive comparison would require access to the source code but not all software is released under open-source; second, even open-source softwares can be difficult to compare, because their requirements in terms of architecture, dependencies etc. are different; finally, not all softwares are well-documented and easy to test in the same way, so non-experienced users may not know all the tweaks to boost simulations.
We compensate the lack of objective experimental comparison with the user feedback provided by this survey.

\section{Survey overview}

The analysis of the survey is reported hereinafter.

\subsection{About the participants}
\begin{figure}
\centering
\includegraphics[width=0.8\hsize]{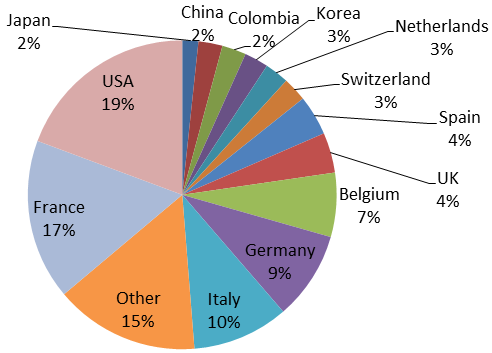}
\caption{Country of provenience for the participants to the survey.}
\label{fig:country}
\end{figure}

The survey was filled by 119 participants (92\% male, 8\% female; age 32 $\pm$ 6, min 20, max 57), whose 62\% holds a PhD degree and 35\% a BS or MS degree, mostly from USA, France, Italy and Germany (see Figure~\ref{fig:country}). Participants work mostly in University (70\%) or do R\&D in public (16\%) or private (14\%) institutes.

Their \textbf{primary areas of research} are:  \\
21\% control, 14\% locomotion, 10\% machine learning, 9\% HRI, 8\% planning, 6\% mechanical design, 5\% cognitive robotics, 5\% mathematical modeling.

Their \textbf{primary application field} is: \\
26\% humanoid robotics, 20\% mobile robotics, 11\% multi-legged robotics, 8\% service robotics, 7\% industrial robotics, 7\% numerical simulation of physical systems, 5\% flying robots. 

Among the participants working in humanoid robotics, 16\% is also competing in the Darpa Robotics Challenge (DRC), which makes 8\% of the participants to the survey  - 10 people.\footnote{Interestingly, the software tool they indicated as the one currently used for their research (we can presume for the DRC as well) is Gazebo (3), MuJoCo (2), Robotran (2), Drake (1), Autolev (1) and ODE (1).}

\subsection{General knowledge about simulating tools}

We asked participants to indicate their familiarity with some of the most common existing simulation tools. We provided a list of existing software tools for simulations, used in different contexts. We asked the users to indicate whether the software was currently used or not for their researches, if it had been used before or if it was unknown. A summary of the percentage of answers for the most relevant tools is shown in Table~\ref{tab:knowledge}.

\begin{table*}
\begin{center}
{\footnotesize
\begin{tabular}{| p{1.1cm}| p{1cm} p{1cm} p{1cm} p{1cm} p{1cm} p{1cm} p{1cm} |}
\hline
Tool  &	Currently used, and it's the main tool	&Currently used, but not the main tool&	Currently used, just to test it&	Used once, just to test it&	Used then abandoned&	Known, but never used&	Never heard of \\
\hline
Gazebo	& \cellcolor{green!25} {\bf 13\%}	& 7\%	& 3\%	& 18\%	& 10\%	& 34\%	& \cellcolor{green!25} {\bf 15\%} \\
ODE	           &\cellcolor{green!25}  {\bf 11\%}	& 12\%	& 5\%	& 18\%	& \cellcolor{red!25}{\bf 22\%}	& 22\%	& \cellcolor{green!25} {\bf 10\%} \\
Bullet	             & 5\%	& 13\%	& 7\%	& 12\%	& 10\%	& 29\%	& 24\% \\
V-Rep	          & 5\%	& 3\%	& 3\%	& 18\%	& 3\%	& 29\%	& 39\% \\
Webots	& 4\%& 	7\%	& 1\%	& 16\%	& \cellcolor{red!25}{\bf 13\%}	& 32\%	& 27\% \\
OpenRave	& 5\%& 	3\%	& 2\%	& 7\%	& 5\%	& 29\%	& 49\% \\
Robotran	& 4\%& 	0\%	& 1\%	& 4\%	& 2\%	& 13\%	& 76\% \\
XDE	& 5\%	& 3\%& 	0\%	& 3\%	& 1\%	& 14\%	& 74\% \\
Blender	& 5\%& 	17\%	& 7\%	& 22\%	& 6\%	& 28\%	&\cellcolor{green!25}  {\bf 15\%} \\
MuJoCo	& 2\%& 	0\%	& 0\%	& 4\%	& 2\%	& 21\%	& 71\% \\
iCub\_SIM	& 4\%	&        4\%	& 2\%	& 3\%	& 3\%	& 29\%	& 55\% \\
Nvidia PhysX	& 1\%& 	1\%	& 4\%	& 12\%	& 7\%	& 43\%	& 32\% \\
OpenSIM	& 3\%	& 4\%	& 3\%	& 8\%	& 1\%	& 41\%	& 40\% \\
HumanS	& 0\%& 	0\%	& 0\%	& 1\%	& 1\%	& 10\%	&\cellcolor{red!25} {\bf 88\%} \\
Moby	         & 2\%	& 1\%	& 0\%	& 0\%	& 2\%	& 14\%	& 81\% \\
Vortex 	& 3\%& 	2\%	& 0\%	& 5\%	& 5\%	& 17\%	& 68\% \\
RoboRobo	& 3\%	& 1\%  	& 0\%	& 0\%	& 1\%	& 4\%& 	\cellcolor{red!25} {\bf 91\%} \\
 \hline
\end{tabular}
}
\end{center}
\caption{Knowledge and past/present use of simulators.}
\label{tab:knowledge}
\end{table*}

The \textbf{software tools that have more than 5\% of user share} (i.e., positive answers to the fact that the software is currently used and it is the one or one of many main tools): the most used are Gazebo (15\%) and ODE (11\%), with a gap with respect to Bullet, OpenRave, V-Rep, XDE and Blender, all at 5\%.
These values provide an indicative dimension of the user community around each software tool.

The software tools that are less known (because maybe they were not sufficiently advertised or do not have a big community behind) and the ones that are most known (even if this does not necessarily means that they are used) can be retrieved from the column ``Never heard of this software'' from Table~\ref{tab:knowledge}\footnote{Actually, Table~\ref{tab:knowledge} is only showing values for the most relevant software tools. To see the full data, we refer the reader to the full report of the survey.}. 
The \textbf{most known tools} are ODE (10\%), Gazebo (15\%), Blender (15\%), Bullet (24\%), Webots (27\%), Nvidia PhysX (32\%), Stage (38\%), V-Rep (39\%), OpenSIM (40\%) and ADAMS (45\%). Interestingly, the first three are also open-source projects.

An important information that we acquired through the survey is about the abandon of software for simulation: this can be found in the column ``Used than abandoned'' in Table~\ref{tab:knowledge}. The\textbf{ most abandoned software after use} are ODE (22\%), Stage (16\%), Webots (13\%), Bullet (10\%), Gazebo (10\%),  Nvidia PhysX (7\%), OpenHRP (6\%), Blender (6\%), OpenRave (5\%), Vortex (5\%). 
Though this set may seem as a sort of ``blacklist'' of tools that disappointed users, it must be observed that most of them are open-source softwares that could have been the ``one among many'' tools that have been used then in one researcher's life; however, it can be equally presumed that the high percentage of abandon can be partly correlated to the difficulty that users have encountered in using these tools and partly by their ``seniority''.

\subsection{Important features for simulation}

We asked participants to indicate the \textbf{main purposes for the use of dynamics simulation in their research} (they could indicate more than one): \\
66\% simulating the interaction of the robot with the environment, 60\% simulating the robot locomotion, 59\% simulating behaviors of the robot before doing them on the real robot, 49\% simulating the robot navigation in the environment, 48\% simulating collisions and interactions between bodies (not specifically robots), 41\% testing low-level controllers for robots, 22\% simulating multi-fingered grasp, 21\% simulating human movements, 8\% animating virtual characters.

We also asked participants to evaluate, upon their experience, what are \textbf{the most important features for a good simulation} (they could evaluate the importance of each element from ``€œnot important at all''€ - 1 to ``€œvery important, crucial''€ -€" 5). Their ranking of important features is reported in Table~\ref{tab:importantfeatures}. The stability of simulation is the only element that was evaluated as ``very important'', whereas speed, precision and accuracy of contact resolution were marked important. Remarkably, the same API between real and simulated robot is also signed as important.

\begin{table*}
\begin{center}
\begin{tabular}{|p{0.4cm}| p{6cm}|l |l|l|}
\hline
Rank&	Feature&	Overall Evaluation & Rating & Median rating \\
\hline
1&	\cellcolor{green!50} \textbf{Stability of simulation} &	Very important & 4.50 $\pm$ 0.58 & 5 \\
2&	\cellcolor{green!25} \textbf{Speed}	&Important & 4.05 $\pm$ 0.75 & 4\\
3&	\cellcolor{green!25} \textbf{Precision of simulation}&	Important & 4.02 $\pm$ 0.71 & 4\\
4&	\cellcolor{green!25}\textbf{Accuracy of contact resolution}	&Important & 3.91 $\pm$ 0.92 & 4\\
5&	\cellcolor{green!25} \textbf{Same interface between real \& simulated system}	&Important & 3.67 $\pm$ 1.26 & 4\\
6&	\cellcolor{yellow!25}Computational load (CPU)	&Neutral & 3.53 $\pm$ 0.85 & 3\\
7&	\cellcolor{yellow!25}Computational load (memory)&	Neutral & 3.22 $\pm$ 0.90 & 3\\
8&	\cellcolor{yellow!25}Visual rendering	&Neutral & 3.02 $\pm$ 1.02 & 3\\
 \hline
\end{tabular}
\end{center}
\caption{Most important features for a simulator.}
\label{tab:importantfeatures}
\end{table*}

\subsection{Criteria for choosing a simulator}

We asked participants to indicate the most important criteria for choosing a simulator. The answer was broken in three parts, i.e. participants could point out the first, second, and third most important criteria.
The first most important criteria: 32\% simulation very close to reality, 24\% open-source, 19\% same code for real and simulated robot, 11\% light and fast, 6\% customization, 3\% no inter-penetration between bodies, 5\% other. 
The second and third choice for the important criteria follow more or less accordingly.
Considering the three criteria as a whole, i.e. grouping the three of them on the same level, the important criteria is 23\% simulation very close to reality, 20\% open-source, 18\% light and fast, 16\% same code for real and simulated robot, 14\% customization, 4\% no-inter-penetration between bodies, 1\% ease to learn/use, 1\% real time -" based simulation, 2\% other. 
If instead we consider the weight of each selection (most important=3, second important=2, third most important=1), then grouping the answers we have: 26\% simulation very close to reality, 22\% open-source, 17\% same code for both real and simulated robot, 17\% light and fast, 11\% customization, 4\% no inter-penetration between bodies (5\% other)

\begin{table}
\begin{center}
\begin{tabular}{|l| l |}
\hline
 Rank & Most important criteria\\
 \hline
1&	Simulation very close to reality\\
2&	Open-source\\
3&	Same code for both real and simulated robot\\
4&	Light and fast\\
5&	Customization\\
6&	No interpenetration between bodies\\
 \hline
\end{tabular}
\end{center}
\caption{Most important criteria for choosing a simulator.}
\label{tab:criteria}
\end{table}

\begin{figure*}
\centering
\includegraphics[width=0.7\hsize]{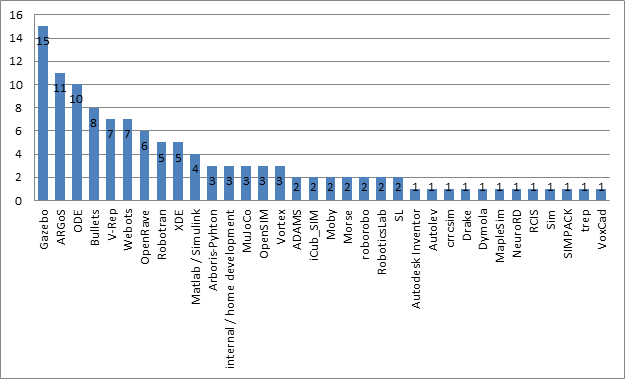}
\caption{The simulation tools currently in use among the participants to the survey. The vertical axis reports the number of  users that indicated the tool as their principal.}
\label{fig:tools}
\end{figure*}

\subsection{Currently used tools}

We asked participants to indicate the current simulation tool they are using. Results are shown in Figure~\ref{fig:tools}.

The \textbf{most diffused software among the participants} are: \\
13\% Gazebo, 9\% ARGoS, 8\% ODE, 7\% Bullet, 6\% V-Rep, 6\% Webots, 5\% OpenRave, 4\% Robotran, 4\% XDE. All the other tools (see Figure 4) have less than 4\% of user share. 

These tools are the ones we are focusing on in our following analysis.
Some technical information about the selected tools can be indicative of the user needs and use:
\begin{itemize}
\item \textbf{Primary OS}: 66\% GNU/Linux, 30\% Windows, 4\% MAC OSX.
\item  \textbf{Primary API language}: 52\% C++, 18\% python, 13\% Matlab, 8\%C, 3\% LUA, 2\% Java; 3\% of participants do not use an API
\item  \textbf{License}: 67\% of the tools are open-source (GPL, Apache, BSD and analogous/derivatives licenses), only 17\% of the tools have a commercial license, 16\% have an academic license (i.e., they are free but not open-source). 
\item  \textbf{Hardware}: 39\% a powerful desktop (i.e., multi-core, 8/16GB RAM), 35\% everyday laptop, 18\% powerful desktop with powerful GPU card, 5\% multi-core cluster.
\item  \textbf{Middleware}: 52\% is not using the tool with a middleware, the remainder is using ROS (25\%), YARP (6\%), OROCOS (4\%).
\end{itemize}

The research areas being different, we extracted the most used tools for a selection of research areas: results are shown in Table~\ref{tab:diffusion}. The most relevant results are for humanoid robotics (31 users, that is 26\% of the participants to the survey) and mobile robotics (25 users, that is 21\% if the participants).
For humanoid robotics, the most diffused tools are ODE and Gazebo, and there is a variety of several custom-made simulators. It is interesting to notice that Gazebo supports ODE and Bullet as physical engines, hence it is probable that the quota of ODE for humanoid robotics is higher. 
For mobile robotics, the most diffused tools among the survey participants are Gazebo, ARGoS and Webots. 

\begin{table*}
\begin{center}
\begin{tabular}{|p{5cm}|p{0.7cm}| p{4.5cm} | p{6cm} |}
\hline
Research area & Users & Most used software & Other used software \\
 \hline
 \hline
 
 Humanoid Robotics &32& (4) ODE, (3) Gazebo, Robotran, OpenRave, Arboris-Python, (2) XDE, iCub\_SIM &  (1) Drake, MapleSim, MuJoCo, OpenSIM, RoboticsLab, SL, Vortex, V-Rep, Webots, own code \\
 \hline
 Mobile Robotics &25& (5) Gazebo, ARGoS, (3) Webots, (2) V-Rep, Vortex &  (1) ADAMS, Autodesk Inventor, Bullet, ODE, Morse, roborobo, Sim, own code\\
 \hline
 Multi-legged robotics &13& (3) Webots, (2) ODE & (1) Gazebo, ADAMS, Autolev, Bullet, Moby, RoboticsLab, SIMPACK, VoxCad \\
 \hline
 Service robotics &12& (4) Gazebo, (3)  OpenRave & (1) OpenSIM, V-Rep, Morse, RCIS, SL \\
 \hline
 Numerical simulation of physical systems&8& (2) Bullet & (1) MuJoCo, ODE, OpenSIM, Simulink, trep, XDE\\
 \hline
 Flying robots &6& (2) ARGoS & (1) Robotran, crrcsim, Gazebo, Simulink/Matlab\\
 \hline
Swarm robotics &5& (4) ARGoS & (1) roborobo\\
\hline
Industrial manipulators &5& & (1) Bullets, Dymola, Matlab, V-Rep, XDE \\
\hline
Mechanical design &4& & (1) Moby, MuJoCo, V-Rep, own code \\
\hline
Human Motion analysis &3& & (1)  Robotran, Bullet, XDE\\
\hline
Snake robots &3&  (2) ODE & (1) Matlab  \\
 \hline
\end{tabular}
\end{center}
\caption{Most diffused tools for a selection of the research areas.}
\label{tab:diffusion}
\end{table*}

The different concentration of tools for the different research areas reveals that some tools are more appropriate than others for simulating robotic systems in different contexts or applications. A researcher may therefore let his choice about the adoption of a simulator be guided by the custom in his field.
With this in mind, we investigated what was the main reason for a researcher to pick up his current tool.
Overall, the main reasons why they chose the current tool is: 29\% the best tool for their research upon evaluation, 23\% ``€œinheritance''€, i.e. it was ``€œthe software''€ (already) used in their laboratory, 8\% they are the developers, 8\% it was chosen by their boss/project leader, 7\% it is open-source, 7\% it was happily used by colleagues. Only 3\% of the participants chose the tool because of a robotic challenge. Interestingly there is quite a demarcation between the first reasons and the others. There are certainly some tools that distinguish for the fact that they have been chosen as best option for research, for example V-Rep (71\%), Bullet (63\%) and Gazebo (53\%). Some tools have instead been adopted by ``inheritance'', i.e., they were already used in the lab: ARGoS (45\%), Robotran (40\%) and XDE (40\%). For the latter, it is also a choice imposed by the project leader (40\%).

We asked participants to evaluate their level of satisfaction of the use of their tool, in a global way, from Very negative (1) to Very Positive (5): all software tools were evaluated ``€œpositive''€, whereas only MuJoCo was ``€œvery positive'' (subjective evaluation by 3 users)€. 
We also asked participants to indicate their level of satisfaction with respect to some specific aspects (documentation, support, installation, tutorials, advanced use, active project and community, API), and to rate each element on a scale from 1 to 5. 
Table~\ref{tab:rankings} reports the mean and standard deviation of the notes received by the users of each tool.

\begin{table*}
\begin{center}
\begin{tabular}{|p{1cm}|p{1.6cm}| p{1.6cm}|p{1.6cm}|p{1.6cm}|p{1.6cm}|p{1.6cm}|p{1.6cm}|p{1.6cm}|}
\hline
Tool	&	Documentation &	Support &	Installation &	 Tutorials &	Advanced use &	Active project \& community &	API	  &	Global \\
\hline
Gazebo	&$ 3.47\pm0.99$ &$ 4.00\pm1.07$ & $3.93\pm1.03$ & $3.53\pm1.12$ & $3.80\pm0.86$ &\cellcolor{green!25} $4.73\pm0.45$ & $3.67\pm0.82$ & $3.88\pm0.91$\\
ARGoS & $3.40\pm0.70$& $3.90\pm0.99$ & $4.70\pm0.48$ & $4.20\pm0.63$ & $4.60\pm0.70$ & $4.10\pm0.74$ & $4.30\pm0.67$ & $4.17\pm0.70$\\
ODE & $3.80\pm0.63$ & $3.40\pm1.07$ & $4.10\pm1.28$ & $3.20\pm1.13$ & $3.90\pm1.37$ & $3.30\pm1.25$ & $3.40\pm1.26$ & $3.59\pm1.15$\\
Bullets & $3.37\pm1.06$ & $3.62\pm0.91$ & \cellcolor{green!25}$4.75\pm0.46$ & $4.00\pm0.76$ & $3.75\pm0.71$ & $4.37\pm0.74$ & $3.87\pm0.83$ & $3.96\pm0.78$\\
V-Rep & \cellcolor{green!25}$4.28\pm0.76$ & \cellcolor{green!25}$4.43\pm0.79$ & $4.71\pm0.76$ & \cellcolor{green!25}$4.14\pm0.90$ & $4.28\pm0.76$ & $4.43\pm0.53$ & $4.14\pm1.07$ &\cellcolor{green!25} $4.25\pm0.80$\\
Webots & $3.86\pm1.07$ & $3.57\pm1.13$ & $4.43\pm0.79$ & $3.43\pm1.51$ & $4.42\pm0.78$ & $4.14\pm0.69$ & $4.57\pm0.53$ & $4.20\pm0.96$\\
OpenRave & $3.50\pm0.55$ & $4.67\pm0.52$ & $4.17\pm0.75$ & $3.50\pm1.22$ & $4.33\pm0.82$ & $4.33\pm0.52$ & $4.33\pm0.52$ & $4.12\pm0.70$\\
Robotran & $3.60\pm0.55$ & $3.80\pm0.45$ & $3.80\pm0.45$ & $3.20\pm0.84$ & $4.20\pm0.84$ & $3.20\pm0.84$ & $3.80\pm0.45$ & $3.66\pm0.63$\\
Vortex & $3.33\pm1.15$ & $3.67\pm1.53$ & $5.00\pm0.00$ & $2.67\pm0.58$ & $3.67\pm0.58$ & \cellcolor{red!25}$2.67\pm1.15$ & $3.33\pm0.58$ & $3.48\pm0.80$\\
OpenSIM & $4.33\pm0.58$ & $4.67\pm0.58$ & $3.67\pm0.58$ & $3.00\pm1.00$ & $4.00\pm0.00$ & $4.67\pm0.58$ & $3.67\pm0.58$ & $4.00\pm0.55$\\
MuJoCo & $2.33\pm1.15$ & \cellcolor{red!25}$1.67\pm0.58$ & $4.33\pm1.15$ & $3.33\pm1.15$ &\cellcolor{green!25} $4.67\pm0.57$ & $4.00\pm0.00$ &\cellcolor{green!25} $5.00\pm0.00$ & $3.62\pm0.66$\\
XDE & \cellcolor{red!25}$1.40\pm0.55$ & $2.80\pm1.09$ & \cellcolor{red!25}$3.60\pm0.55$ & \cellcolor{red!25}$2.80\pm1.09$ &\cellcolor{red!25} $3.40\pm1.10$ & $2.80\pm0.84$ & \cellcolor{red!25}$3.00\pm1.00$ & \cellcolor{red!25}$2.83\pm1.07$\\
\hline
\end{tabular}
\end{center}
\caption{Ratings for the level of user satisfaction of the most diffused tools.}
\label{tab:rankings}
\end{table*}

\begin{figure*}
\centering
\includegraphics[width=0.7\hsize]{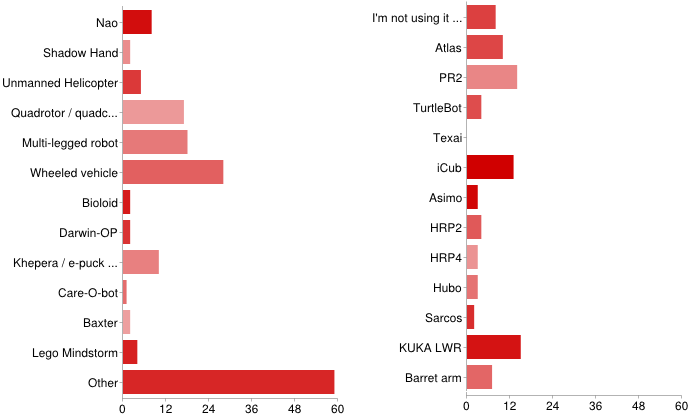}
\caption{The simulation tools currently in use among the participants to the survey. The vertical axis reports the number of  users that indicated the tool as their principal.}
\label{fig:robots}
\end{figure*}

\subsection{Tools for robots}

The majority of participants to the survey is using the software tool to simulate robots (91\%). Users could point out the robots they are simulating (more than one in general): the aggregated table of simulated robot is shown in Figure~\ref{fig:robots}, where the x-axis shows how many users selected the robot.

We extracted the principal tools used for simulating the main robots:
\begin{itemize}
\item iCub: 25\% Arboris-Python, 17\% ODE, 17\% Robotran, 17\% iCub\_SIM
\item Atlas: 50\% Gazebo, 25\% MuJoCo, 12\% Autolev, 12\% Drake
\item PR2: 21\% OpenRave, 14\% Gazebo, 14\% MuJoCo, 7\% Bullet, 7\% V-Rep
\item Multi-legged robot: 22\% ODE, 11\% SL, 11\% Bullet, 11\% Webots
\item Wheeled vehicle: 14\% Gazebo, 14\% V-Rep, 11\% ARGoS, 7\% Morse, 7\% Webots, 7\% Vortex
\item Quadrotor: 24\% Gazebo, 24\% ARGoS, 12\% V-Rep
\end{itemize}


\section{Software information fiches}

We report in the following some essential information for the main software tools (the most diffused) that may be of help for the interested reader. Most of the information gathered here is extracted from the survey (each item is marked by a filled dot, {\footnotesize $\bullet$}). When it is not the case, an empty dot $\circ$ is used.
For the subjective user feedback we refer the reader to the full report of the survey. Data are reported with $\%$, however to have a fair comparison we report in brackets the number of participants that selected the specified tool.
Note that in the following ``main simulated robots'' refers to real robots that are simulated in the software.

\subsection{Gazebo}

Gazebo is a multi-robot simulator for outdoor environments, developed by Open-Source Robotics Foundation. It is the official software tool for the DRC. It supports multiple physics engines (ODE, Bullet).
\begin{itemize}
\item[$\circ$] Web: \url{http://gazebosim.org/}
\item[$\circ$] License: Apache 2
\item Survey participants: 15
\item	OS share: 100\% GNU/Linux
\item		Main API: 80\% C++
\item		Main reason for adoption: 53\% best tool upon evaluation, 20\% software already used in the lab, 20\% official tool for a challenge, 7\% open-source 
\item 	Mostly used in USA (33\%)
\item		Mainly used for: 33\% mobile robotics, 27\% service robotics, 20\% humanoid robotics
\item             Main simulated robots: 40\% Atlas, 33\% custom platform, 27\% wheeled vehicle, 27\% quadrotor, 27\% turtlebot, 20\% PR2
\item             Main middleware used with: 93\% ROS
\item   Main simulated robots: 40\% Atlas, 33\% custom platform, 27\% wheeled vehicle, 27\% quadrotor, 27\% turtlebot, 20\% PR2
\end{itemize}

\subsection{ARGoS}

ARGoS is a multi-robot, multi-engine simulator for swarm robotics, initially developed within the Swarmanoid project\footnote{\url{http://www.swarmanoid.org/}}.
\begin{itemize}
\item[$\circ$] Web: \url{http://iridia.ulb.ac.be/argos/}
\item[$\circ$] License: GPLv3.0
\item Survey participants: 11
\item	OS share: 91\% GNU/Linux, 9\% MAC OSX
\item		Main API: 73\% C++
\item		Main reason: 45\% software already used in the lab, 27\% colleagues using it
\item		Mostly used in Belgium (36\%) and Italy (27\%)
\item		Used for: 46\% mobile robotics, 36\% swarm robotics, 18\% flying robots
\item  Main simulated robots: 64\% khepera/e-puck/thymio, 36\% marXbot/footbot, 27\% quadrotor

\end{itemize}

\subsection{ODE}

ODE (Open Dynamics Engine) is an open-source library for simulating rigid body dynamics, used in many computer games and simulation tools. It is used as physics engines in several robotics simulators, such as Gazebo and V-Rep.
\begin{itemize}
\item[$\circ$]  Web: \url{http://www.ode.org/}
\item[$\circ$]  License: GNU LGPL and BSD
\item Survey participants: 10
\item	OS share: 100\% GNU/Linux
\item	Main API: 80\% C++
\item	Main reason: 50\% best tool upon evaluation, 20\% used before, 10\% boss choice, 10\% open-source, 10\% software already used in the lab
\item	Mostly used in France (20\%)
\item	Used for: 50\% humanoid robotics, 20\% multi-legged robotics, 20\% snake robots, 10\% numerical simulation of physical systems
\item Main simulated robots: 40\% multi-legged robot, 20\% iCub
\end{itemize}

\subsection{Bullet}
Bullet is an open-source physics library, mostly used for computer graphics and animation. The latest release\footnote{At the time we are submitting this paper, the latest version is 2.82, released at the end of october 2013 - after the survey.} also supports Featherstone's articulated body algorithm and a Mixed Linear Complementarity Problem solver, which makes it suitable for robotics applications.
\begin{itemize}
\item[$\circ$] Web: \url{http://bulletphysics.org}
\item[$\circ$] License: ZLib license, free for commercial use
\item Survey participants: 8
\item	OS share: 50\% Windows, 38\% GNU/Linux, 12\% MAC OSX
\item	Main API: 75\% C++
\item	Main reason: 63\% best tool upon evaluation, 25\% open-source, 12\% colleagues using it
\item	Mostly used in France (25\%), Italy (25\%) and Belgium (25\%)
\item	Used for: 25\% humanoid robotics, 25\% numerical simulation of physical systems, 12.5\% industrial manipulators, 12.5\% human motion analysis, 12.5\% mobile robotics, 12.5\% multi-legged robotics
\item   Main simulated robots: 25\% multi-legged robot
\end{itemize}

\subsection{V-Rep}
V-Rep is a robot simulator software with an integrated development environment, produced by Coppelia Robotics. Like Gazebo, it supports multiple physics engines (ODE, Bullet, Vortex).
\begin{itemize}
\item[$\circ$] Web: \url{http://www.coppeliarobotics.com/}
\item[$\circ$] License: Dual-licensed source code: commercial or GNU GPL
\item Survey participants: 7
\item	OS share: 57\% GNU/Linux, 43\% Windows
\item	Main API: 57\% C++, 29\% LUA
\item  Middleware: 43\% ROS, 57\% None
\item	Main reason: 72\% best tool upon evaluation, 14\% colleagues using it, 14\% boss choice
\item	Used for: 29\% mobile robotics, 14\% industrial manipulators, 14\% humanoid robotics, 14\% mechanical design, 14\% cognitive architectures, 14\% service robotics
\item	Main simulated robots: 29\% Nao, 29\% quadrotor, 29\% wheeled vehicle, 29\% Bioloid, 29\% khepera/ e-puck/ thymio
\end{itemize}

\subsection{Webots}
Webots is a development environment used to model, program and simulate mobile robots developed by Cyberbotics Ltd.
\begin{itemize}
\item[$\circ$] Web:  \url{http://www.cyberbotics.com}
\item[$\circ$] License:  Commercial or limited features free academic license
\item Survey participants: 7
\item	OS share: 57\% GNU/Linux, 29\% Windows, 14\% MAC OSX
\item	Main API: 71\% C++
\item	Main reason: 29\% best tool upon evaluation, 29\% software already used in the lab, 14\% boss choice, 14\% official tool for a challenge, 14\% used before
\item	Used for: 43\% mobile robotics, 43\% multi-legged robotics, 14\% humanoid robotics
\item Main simulated robots: 29\% KUKA LWR, 29\% Lego Mindstorm, 29\% wheeled vehicle
\end{itemize}

\subsection{OpenRave}

OpenRave is an environment for simulating motion planning algorithms for robotics.
\begin{itemize}
\item[$\circ$] Web: \url{http://openrave.org/}
\item[$\circ$] License: LGPL and Apache 2
\item Survey participants: 6
\item		OS share: 100\% GNU/Linux
\item		Main API: 83\% python
\item		Main reason: 50\% best tool upon evaluation, 33\% colleagues using it, 17\% boss choice
\item		Mostly used in USA (33\%)
\item		Used for: 50\% humanoid robotics, 50\% service robotics
\item Main simulated robots: 50\% PR2
\end{itemize}

\subsection{Robotran}

Robotran is a software that generates symbolic models of multi-body systems, which can be analysed and simulated in Matlab and Simulink.
It is developed by the Center for Research in Mechatronics, Universit\'e Catholique de Louvain.
\begin{itemize}
\item[$\circ$] Web: \url{http://www.robotran.be/}
\item[$\circ$] License: commercial and free non commercial license
\item Survey participants: 5
\item		OS share: 80\% Windows, 20\% GNU/Linux
\item		Main API: 60\% C
\item		Main reason: 40\% software already used in the lab, 20\% best tool upon evaluation, 20\% developer, 20\% open-source (free)
\item		Used only in Belgium (40\%) and Italy (60\%)
\item		Used for: 60\% humanoid robotics, 20\% human motion analysis, 20\% flying robots
\item Main simulated robots: 60\% Coman, 40\% iCub
\end{itemize}

\subsection{XDE}
XDE is an interactive physics simulation software environment fully developed by CEA LIST.
\begin{itemize}
\item[$\circ$] Web: \\
\url{http://www.kalisteo.fr/lsi/en/aucune/a-propos-de-xde}
\item[$\circ$] License:	Commercial and free non commercial license
\item Survey participants: 5
\item		OS share: 60\% GNU/Linux, 40\% Windows
\item		Main API: 100\% python
\item[$\circ$]     Middleware: OROCOS
\item		Main reason: 40\% boss choice, 40\% software already used in the lab, 20\% developer
\item		Used only in France (100\%)
\item		Used for: 40\% humanoid robotics, 20\% industrial manipulators, 20\% numerical simulation of physical systems, 20\% human motion analysis
\item Main simulated robots: 40\% industrial robots, 40\% KUKA LWR, 20\% iCub, 20\% wheeled vehicle
\end{itemize}

\section{Conclusions}

With the growing interest of robotics for physical interaction, simulation is no longer a tool for offline computation and visualization, but is used in particular for rapidly prototyping controllers. That is why researchers stressed the importance of more realistic simulation, same code for both real and simulated robot, beside the availability of the source code.

This shift in the expectations from simulation reflects in the migration from physics engines classically used for animation of virtual characters and computer graphics towards physics engines supporting  robotics descriptions of bodies and more contact solvers.  
The users' knowledge of multiple simulation tools and their activity in testing and abandoning eventually a tool, suggest that users look for the right tool that meets their requirements and is fit for their problem.
For instance, the robotics community demands physics engines with direct support of robotics descriptions of multi-body systems. This is the reason why Bullet is now supporting LCP solvers and Featherstone's ABA, and new physics engines like MuJoCo\footnote{MuJoCo is not merely a physics engine, it incorporates control and optimization modules.} or Vortex have been created.

A good compromise is a modular software that supports multiple physical engines, enabling a tradeoff between simulation accuracy and computational resources. Those features, together with the stability of the simulation, are of main concern for the users.
This strategy, adopted with Gazebo by the research community and with V-Rep at industrial level, seems to pay off in terms of user feedback, because the first is the most diffused among the survey participants and the second the best rated. Subjective free-comments\footnote{They can be read in the extended version of the survey report: \REPORTARXIV.} reported that users of those tools, though acknowledging their current limitations, were confident in the announced developments that could sensibly improve the tools.

To conclude, we overviewed the panorama of simulation tools that are currently used in robotics.
Each software inherits its specificities from the expected domains of application or the original application for which is was conceived, which results in a variety of tools with different features ranging from dynamic solver libraries to systems simulation software. 
More recent tools, like Gazebo and V-Rep, have the potential to be of general use thanks to their good support and community and the support of different physical engines.
Notwithstanding, we remind that designing a perfect physics engine is impossible and there will always be a difference between simulation and reality, a gap that should be taken into account by the simulator and the robot controllers~\cite{Mouret2012}.

\begin{figure*}
\begin{center}
\begin{subfigure}[b]{0.5\textwidth}
\centering
\includegraphics[height=5cm]{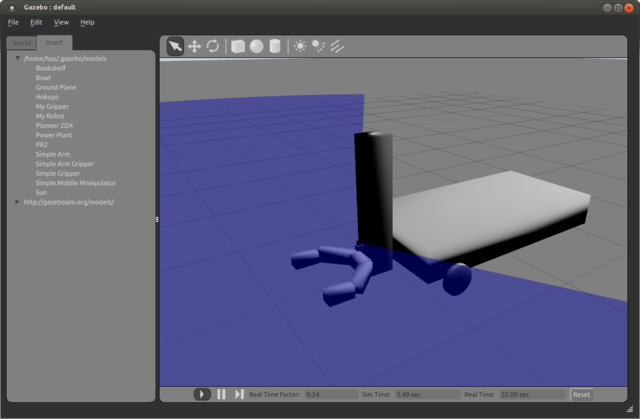} 
\caption{Gazebo.}
 \label{fig:xde}
 \end{subfigure}
\begin{subfigure}[b]{0.45\textwidth}
\centering
\includegraphics[height=5cm]{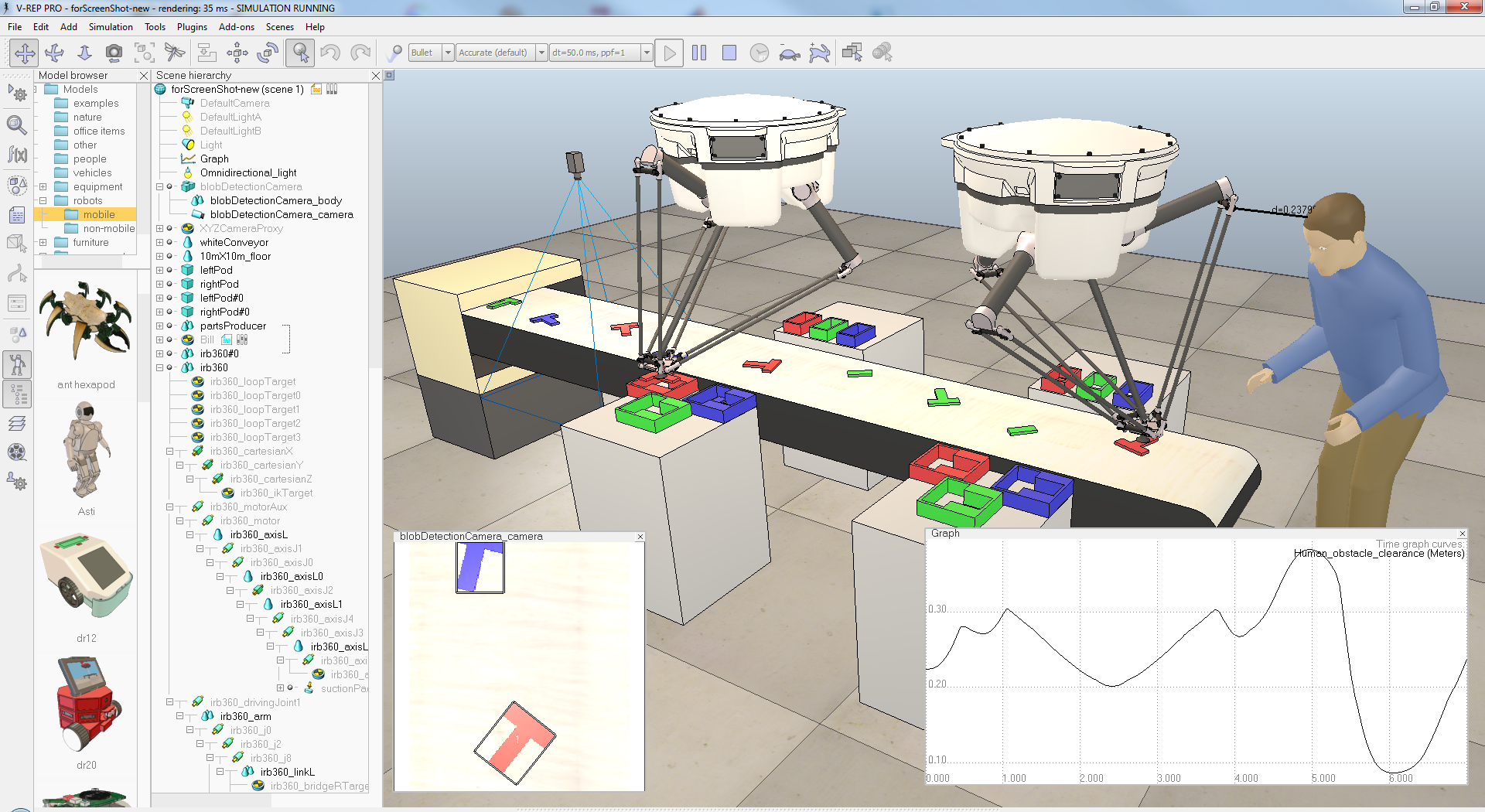}
\caption{V-Rep.}
 \label{fig:vrep}
 \end{subfigure}

\caption{The simulation environment of Gazebo and V-Rep (credits:  \texttt{http://gazebosim.org} and \texttt{http://www.coppeliarobotics.com}).}
\label{fig:simulators}
\end{center}
\end{figure*}

\begin{figure*}
\begin{center}
\begin{subfigure}[b]{0.56\textwidth}
\centering
\includegraphics[height=6.5cm]{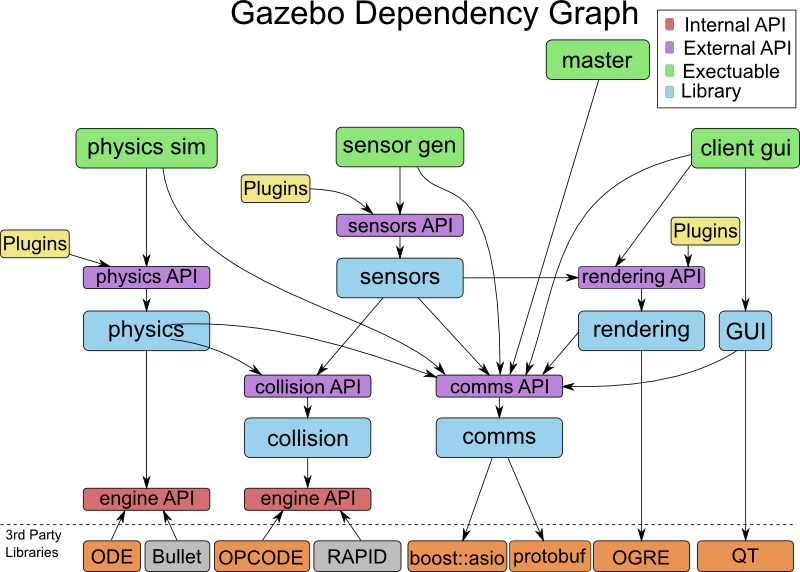} 
\caption{Gazebo}
 \label{fig:xde}
 \end{subfigure}
\begin{subfigure}[b]{0.4\textwidth}
\centering
\includegraphics[height=6.5cm]{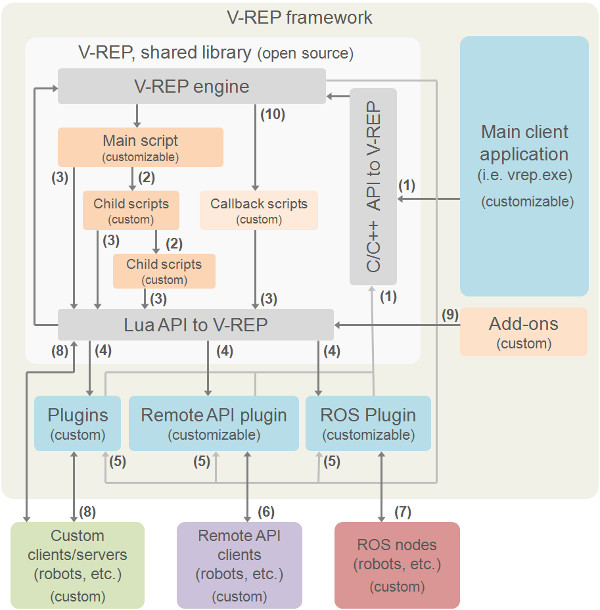}
\caption{V-Rep}
 \label{fig:vrep}
 \end{subfigure}

\caption{A graphical representation of the software architectures of Gazebo and V-Rep (credits: \texttt{http://gazebosim.org} and \texttt{http://www.coppeliarobotics.com}).}
\label{fig:simulators}
\end{center}
\end{figure*}

\section*{Acknowledgment}

The authors are supported by the EU Project CODYCO (FP7-ICT-2011-9, No. 600716) - \url{www.codyco.eu}.

\ifCLASSOPTIONcaptionsoff
  \newpage
\fi

\bibliographystyle{IEEEtran}
\bibliography{IEEEabrv,codycosimulators}
%
%
%

%

\begin{IEEEbiography}[{\includegraphics[width=1in,height=1.25in,clip,keepaspectratio]{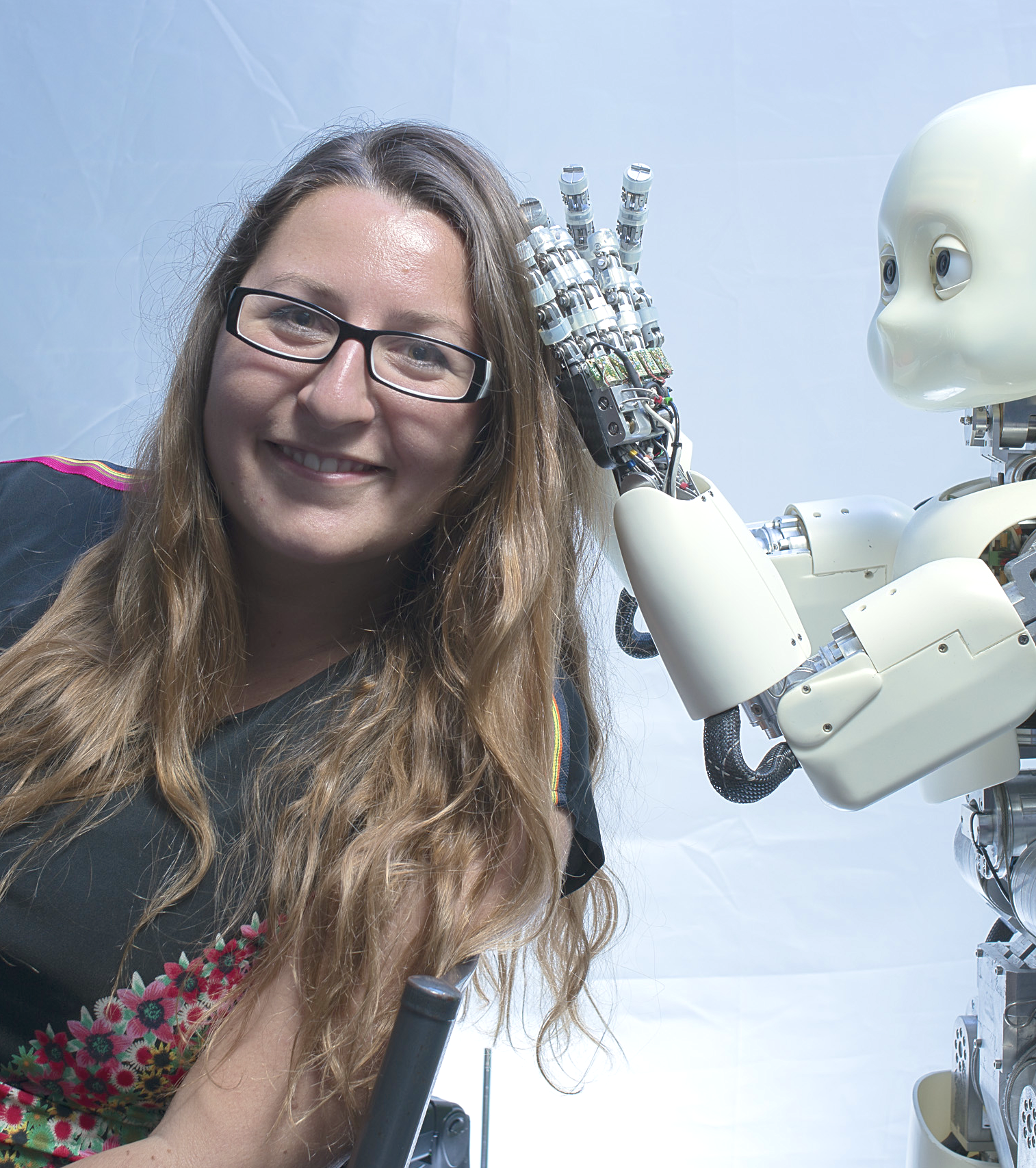}}]{Serena Ivaldi}
received the M.S. degree in Computer Engineering with highest honors in 2006 at the University of Genoa (Italy) and her PhD in Humanoid Technologies in 2011, jointly at the University of Genoa and Italian Institute of Technology. There she also held a research fellowship in the Robotics, Brain and Cognitive Sciences Department. Since 2011 she is a postdoctoral researcher in the Institut des Syst\`emes Intelligents et de Robotique (ISIR), where she coordinates the experiments of MACSi, EDHHI and CODYCO projects on iCub.  Her research is centered on humanoid robots interacting physically with humans and environment.
Web: http://chronos.isir.upmc.fr/~ivaldi

\end{IEEEbiography}

\begin{IEEEbiography}[{\includegraphics[width=1in,height=1.25in,clip,keepaspectratio]{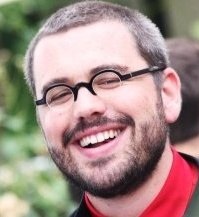}}]{Vincent Padois}
is an associate professor of Robotics and Computer Science and a member of the Institut des Syst\`emes Intelligents et de Robotique (ISIR, UMR CNRS 7222) at Universit\'e Pierre et Marie Curie (UPMC) in Paris, France. In 2001, he receives both an engineering degree from the Ecole Nationale d'Ing\`enieurs de Tarbes (ENIT), France and his master degree in Automatic Control from the Institut National Polytechnique de Toulouse (INPT), France. From 2001 to 2005, he is a PhD student in Robotics of the ENIT/INPT Laboratoire G\'enie de Production. In 2006 and 2007, he is a post-doctoral fellow in the Stanford Artificial Intelligence Laboratory  and more specifically in the group of Professor O. Khatib. Since 2007, his research activities at ISIR are mainly focused on the automatic design, the modelling and the control of redundant and complex systems such as wheeled mobile manipulators, humanoid robots as well as standard manipulators evolving under constraints in complex environments. He is also involved in research activities that aim at bridging the gap between adaptation and decision making techniques provided by Artificial Intelligence and low-level, reactive control. Since 2011, he holds the "Intervention Robotics" RTE/UPMC chair position.
\end{IEEEbiography}


\begin{IEEEbiography}[{\includegraphics[width=1in,height=1.25in,clip,keepaspectratio]{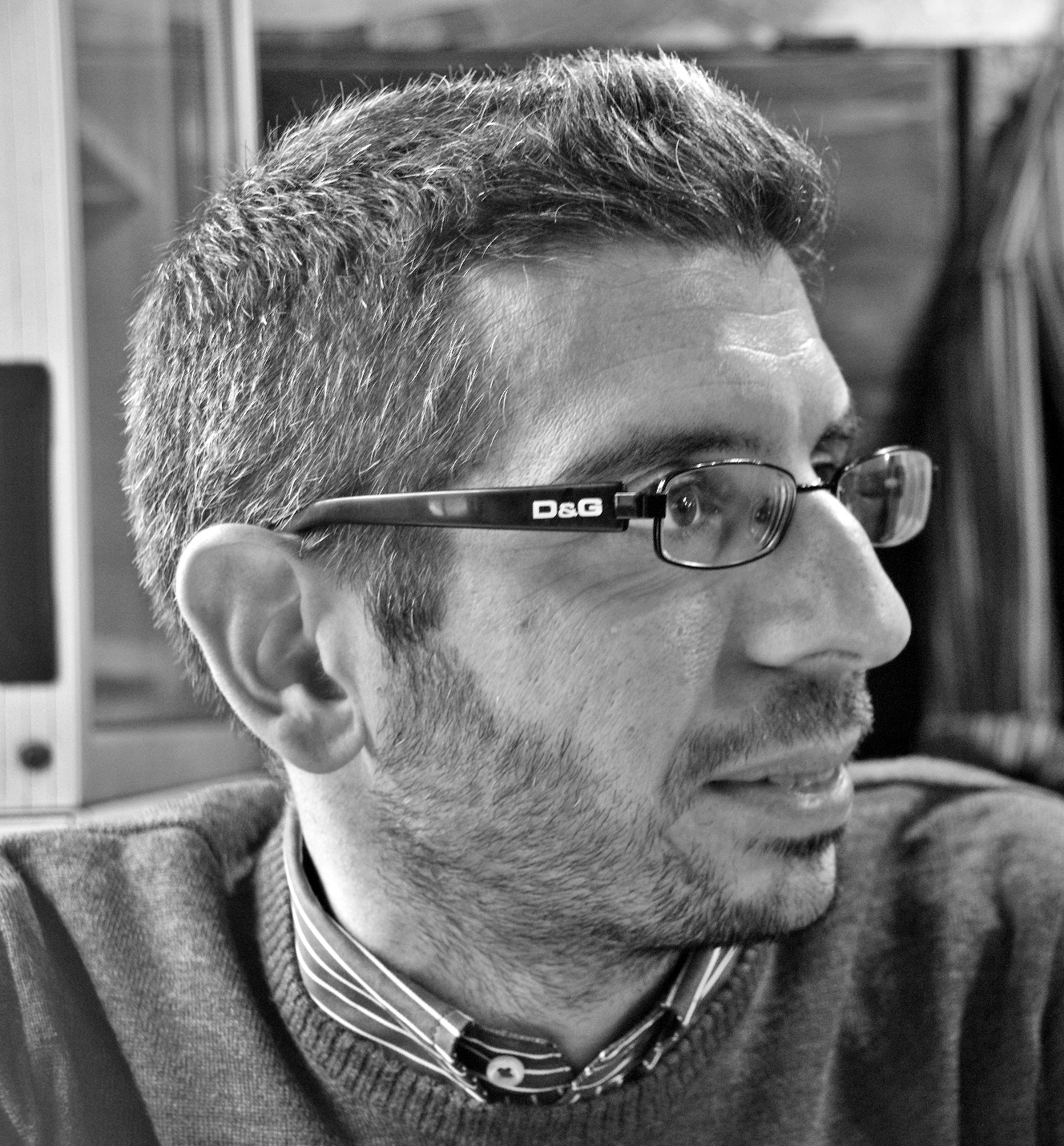}}]{Francesco Nori}
 was born in Padova in 1976. He received his D.Ing. degree (highest honors) from the University of Padova (Italy) in 2002. He received his Ph.D. in Control and Dynamical Systems from the University of Padova (Italy) in 2005. From 2007 he joined the Istituto Italiano di Tecnologia, contributing significantly to the development of the iCub humanoid robot. His research interest are currently focused on whole-body motion control exploiting multiple contacts.
\end{IEEEbiography}




\appendices

\section{Users knowledge of simulation tools}

We asked subjects to indicate their familiarity with some existing simulation tools. We provided a list of existing software tools for simulations, used in different contexts. We asked the users to indicate whether the software was currently used or not for their researches, if it had been used before or if it was unknown. 
A summary of the percentage of answers for the most relevant tools was shown in Table~\ref{tab:knowledge}.

The \textbf{most currently used main tools} (i.e., tools that have more than 5\% of positive answers to the fact that the software is currently used and it is the main tool) are reported in Table~\ref{tab:bestSW}. The \textbf{least unknown software tools} (i.e., tools that were marked as ``never heard of'' by the users) are reported in Table~\ref{tab:leastknownSW}.
The software tools that have been abandoned the most after use are reported in Table~\ref{tab:mostabandonedSW}.

\begin{table}
\begin{center}
\begin{tabular}{|l|l|l|}
\hline
Rank&	Most currently used main tool ($>$5\%) & \% user share \\
\hline
1&	Gazebo & 13\% \\
2&	ODE & 11\% \\
3&      Bullet & 5 \% \\
4&	OpenRave&	5\%\\
5&	V-Rep&	5\%\\
6&	XDE&	5\%\\
7&	Blender&	5\%\\
 \hline
\end{tabular}
\end{center}
\caption{Best software upon user rating.}
\label{tab:bestSW}
\end{table}

\begin{table}
\begin{center}
\begin{tabular}{|l|l|p{2cm}|}
\hline
Rank&	Least unknown software tool & \% user that have never heard about this software\\
\hline
1&	ODE&	10\%\\
2&	Gazebo&	15\%\\
3&	Blender&	15\%\\
4&	Bulle&t	24\%\\
5&	Webots&	27\%\\
6&	Nvidia PhysX&	32\%\\
7&	Stage&	38\%\\
8&	V-Rep&	39\%\\
9&	OpenSIM&	40\%\\
10&	ADAMS&	45\%\\
 \hline
\end{tabular}
\end{center}
\caption{Least unknown software tool.}
\label{tab:leastknownSW}
\end{table}

\begin{table}
\begin{center}
\begin{tabular}{|l|l|p{2cm}|}
\hline
Rank&Most abandoned software tool & \% user that abandoned this software\\
\hline
1&	ODE&	22\%\\
2&	Stage&	16\%\\
3&	Webots&	13\%\\
4&	Bullet&	10\%\\
5&	Gazebo&	10\%\\
6&	Nvidia PhysX&	7\%\\
7&	OpenHRP&	6\%\\
8&	Blender&	6\%\\
9&	OpenRave&	5\%\\
10&	Vortex&	5\%\\

 \hline
\end{tabular}
\end{center}
\caption{Most abandoned software tool.}
\label{tab:mostabandonedSW}
\end{table}

\section{Free comments about major problems in simulation}

We report hereby the users free comments about their selected simulation tool. We choose to not alter the answers (e.g., correct grammar, punctuation, etc.) to preserve the integrity of the user's answers, except for bad language that was replaced by ***.

\begin{itemize}

\item \textbf{ADAMS}
\begin{itemize}
\item Adams could be too slow and requires a lot of system resources.
\item Speed Accuracy
\end{itemize}

\item \textbf{Arboris-Python}
\begin{itemize}
\item Lack of surface/surface contact models Use of primitive shapes only
\item Arboris-Python is slow and has a limited set of features.
\item Computation speed is slow.
\end{itemize}

\item \textbf{ARGoS}
\begin{itemize}
\item In general, I am very happy with ARGoS, because it does exactly what I expect it to do.
\item a bit slow for swarming
\item Limited to the computational capability of the computer which is provided by the school.
\item more documentation C++ ***
\item Difference between reality and simulation (sensors)
\item Issue when gripping objects.
\item None
\item Visualization of additional information related to the simulation.
\item Better debugging facilities would be nice.
\item for now none.
\end{itemize}

\item \textbf{Autodesk Inventor}
\begin{itemize}
\item Programmin for that is not user-friendly.
\end{itemize}

\item \textbf{Autolev}
\begin{itemize}
\item Impact and contact modeling.
\end{itemize}

\item \textbf{Bullet}
\begin{itemize}
\item The simulation of floor contact. I need very high friction contact points that don't slide on the floor and this doesn't work quite right.
\item Not precise enough
\item Does not apply to my case
\item Lack of continuos collision detection
\item Lack of a well understood and properly calibrated contact model.
\item Not able to handle kinematic chains well. Oscillation of objects.
\item Realism and precision
\end{itemize}

\item \textbf{crrcsim}
\begin{itemize}
\item customization is done in c++, code is quite convoluted
\end{itemize}

\item \textbf{Drake}
\begin{itemize}
\item The fact that I have to write and maintain it myself.
\end{itemize}

\item \textbf{Dymola}
\begin{itemize}
\item being physics based, usually realistic simulations need a very deep knowledge of underlying physical parameters (e.g. for contact)
\end{itemize}

\item \textbf{Gazebo}
\begin{itemize}
\item Lack of documentation, quite slow when simulation start to be a bit complex. Collisions may results in non-realistic jumps of the robot.
\item The simulations are very slow if it is in a real environment.
\item Gazebo does not yet support all Bullet features, especially non-rigid bodies although we will need this in future - that might lead to abandance of Gazebo and use Bullet directly, although lots of efforts
\item Customization
\item The dynamics engine is not very sophisticated
\item Tuning simulated pid controllers for having stable simulation when real inertias and robot model are used.
\item It's hard to create worlds.
\item Real-time factor is $>$50\%. Makes it very difficult to use in semi-virtual experimental setups, where simulated environment is a part of human-centric system.
\item Fast simulation required (1kHz+) AND good contacts/frictions (this is e.g. when using XDE, not with Gazebo...)
\item high computational load $->$ does not cause failures, but costs lots of time
\item With all simulation tools I have tried so far, just installing and setting up the simulator properly has been very time and effort consuming.
\item Lacking ability to run it on multiple operating systems. Lacks rewindability and very slow playback of logs.
\item Deformation modeling
\item Customization of large environment is time consuming.
\item Simulation of contact with non-rigid bodies like terrain. Difficulty in simulating fast and dynamic motions. Computationally slow and demanding.
\end{itemize}

\item \textbf{own code}
\begin{itemize}
\item Not completely realistic simulation of fast humanoïd motions, due to the fact an internal ankle flexibility was represented by the compliant ground contact
\end{itemize}

\item \textbf{iCub\_SIM}
\begin{itemize}
\item No force sensors available on the simulator.
\item Can run slowly.
\end{itemize}

\item \textbf{MapleSim}
\begin{itemize}
\item No major problem: my own code is optimized for speed, which is crucial for numerical optimization.
\end{itemize}

\item \textbf{Matlab}
\begin{itemize}
\item They do not model friction well enough
\item Yet it is fast performing a complete simulation cycle, It is not close to real-time, which is a disadvantage because re-setup simulation parameters can not be don "on the fly". The accuracy of the simulator, depends of prior validations and tuning. This make some tests slow. This lack of robustness, can be improved in the future
\end{itemize}

\item \textbf{Moby}
\begin{itemize}
\item Trying to make sure that our control loops run at the same rate on the simulation as they do on our hardware.
\item I'm not having any problems with the simulations.
\end{itemize}

\item \textbf{Morse}
\begin{itemize}
\item integration with other parts of the overall robotics ecosystem. In other words: customization, reuse, composability
\item setting up dependencies of software, generating simulated environments as i am not familiar with blender
\end{itemize}

\item \textbf{MuJoCo}
\begin{itemize}
\item Meshes colors are not supported yet
\item Modeling accuracy. What you call "Gap between simulation and reality"
\item Visualization
\end{itemize}

\item \textbf{NeuroRD}
\begin{itemize}
\item They take a long time to complte.
\end{itemize}

\item \textbf{ODE}
\begin{itemize}
\item - lack of feedback by the dynamic engine (what is the torque actually applied at this joint, after having applied all the constraints?) - lack of stability - no way to predict how confident the simulator is
\item CPU load
\item Yet it is fast performing a complete simulation cycle, It is not close to real-time, which is a disadvantage because re-setup simulation parameters can not be don "on the fly". The accuracy of the simulator, depends of prior validations and tuning. This make some tests slow. This lack of robustness, can be improved in the future.
\item Lack of documentations, steep learning curve, and hard customization.
\item Yet it is fast performing a complete simulation cycle, It is not close to real-time, which is a disadvantage because re-setup simulation parameters can not be don "on the fly". The accuracy of the simulator, depends of prior validations and tuning. This make some tests slow. This lack of robustness, can be improved in the future.
\item No robot-model file-formats are supported (e.g. COLLADA, VRML, etc.)
\item stability and numerical accuracy to handle stiff contacts
\item Computationally demanding for simulating full 53 degrees of freedom. In particular simulating collisions between small rigid bodies such as the iCub's hands.
\item The contact forces that I need to validate my controller are often unrealistic.
\item 1. Stability is the most problem
\end{itemize}

\item \textbf{OpenRave}
\begin{itemize}
\item Slow
\item Simulating compliant surfaces is difficult/poorly supported.
\item Unstable physics, even can change upon release.
\item The Collision detection module is extremely computational expensive for non-convex objects,so the usage is limited only for convex objects if the simulations needs to run in reasonable time.
\item Better friction/contact rendering
\end{itemize}

\item \textbf{OpenSIM}
\begin{itemize}
\item Sometimes they are really hard to design.
\item Generating new models of robots or musculoskeletal systems.
\item lack of real-time representation
\end{itemize}

\item \textbf{RCIS}
\begin{itemize}
\item The simulation is computationally demanding if you need to run several instances of the simulation in parallel.
\end{itemize}

\item \textbf{roborobo}
\begin{itemize}
\item The fastest the better
\item It takes some knowledge of the simulator to set up the simulation. Unless having one common api for all simulators I don't think that this problem can be solved
\end{itemize}

\item \textbf{RoboticsLab}
\begin{itemize}
\item Collision is unstable very occasionally but it can be fixed by adjusting parameters.
\item C++ programming environment.
\end{itemize}

\item \textbf{Robotran}
\begin{itemize}
\item Numerical instability when a control input is non-smooth or if the model is very stiff.
\item The user has to compute the contact forces and the collision detection
\item Identification of the human physiology
\item Speed
\end{itemize}

\item \textbf{SIMPACK}
\begin{itemize}
\item fast
\end{itemize}

\item \textbf{Simulink + spatial\_v2}
\begin{itemize}
\item slow (because of Matlab)
\end{itemize}

\item \textbf{Simulink and Matlab}
\begin{itemize}
\item it takes a lot of time to build what is not exists.
\end{itemize}

\item \textbf{SL}
\begin{itemize}
\item In case of SL the biggest issue could be collision, since the robot geometry is not considered. Also, only point contact is considered. This could be improved.
\item Documentation
\end{itemize}

\item \textbf{trep}
\begin{itemize}
\item Our software has no automatic handling of contact or impact maps. Additionally, adaptive time-stepping breaks some of the guarantees of structured integration.
\end{itemize}

\item \textbf{Vortex}
\begin{itemize}
\item Lack of the proper documentation, simulation of sand and digging in it.
\item not specific to software: getting material properties and other parameters right
\item Having to write copious amounts of C++ to get things going.
\end{itemize}

\item \textbf{VoxCad}
\begin{itemize}
\item Need more speed
\end{itemize}

\item \textbf{V-Rep}
\begin{itemize}
\item They can be slower on an older laptop, and we need real time(ish) actuation when the operator is in the loop. But that is just a question of horse power.
\item repetability of some dynamic events (e.g. grasping when fingered)
\item GPU and redu function not vast enough
\item No problems until now. It does what I need it to do.
\item Most of them take place in the future ;-) meaning: the above answers are based on what I want to do with V-Rep, but I have not yet found the time and resources to do it.
\item No elastic bodies simulation support
\item Difficulty building models for multi-point grasping (humanoid style hand).
\end{itemize}

\item \textbf{Webots}
\begin{itemize}
\item difficulty in describing dynamics.
\item Slowness: for evolutionary robotics, I need many repeats/trials. When simulating modular robots, the system grinds to a halt when simulating more than a couple of dozen modules.
\item our controller require the simulation to be realtime, which it is not on complicated environments when the camera is simulated. The root problem is our controller not the simulator though.
\item Inaccuracies, contact model, MODELING COMPLIANCE IN A GOOD WAY WITHOUT SIMULATION EXPLOSION, closed chain kinematics, being different from reality, difficulty porting to the robot
\item robot planning algorithms
\item Robot description could be made more user friendly
\end{itemize}

\item \textbf{XDE}
\begin{itemize}
\item Compatibilty with software dependencies
\item Documentation may be insufficient to solve some difficulties in the setup/control of the simulation.
\item Customizable tradeoff between simulation precision and computational resources
\item Dynamic of human movements is to fast to be accuratley reproduced in simulation : movemenst are not the same and fast movements leads to a lost of balance and fall of the manikin.
\item contacts forces time evolution
\end{itemize}

\end{itemize}

\section{Free comments about desirable features missing in the software}

\begin{itemize}

\item \textbf{ADAMS}
\begin{itemize}
\item Easy interface with pro/e or other CAD softwares.
\item customization ( coding a new body element) "Rethinking the modeling the technique"
\end{itemize}

\item \textbf{Arboris-Python}
\begin{itemize}
\item Catalogue of contact models Use of custom shapes
\item C++ coding, general collision detection.
\item It lacks of the same code for real and simulating environment.
\end{itemize}

\item \textbf{ARGoS}
\begin{itemize}
\item Integration with other tools is the next big task to work on. Also, increasing the number of supported robots.
\item a nice documentation, but as the support is reactive, it compensiates.
\item Changing the environment without having to pause the simulation and moving the objects manually.
\item better logging infrastructure
\item Magnetics and soft bodies + contact forces simulation
\item I'm sure many, but for my work, none.
\item None
\item Visualization: - showing additional metadata for each robot (current values of variables etc.); - interactive exploration of the current situation (e.g. shortest path visualizations).
\item More robots simulated.
\item scripting, but it's coming soon
\end{itemize}

\item \textbf{Autodesk Inventor}
\begin{itemize}
\item A toolbox in Matlab or Simulink for automated connection between Inventor and Simulink
\end{itemize}

\item \textbf{Autolev}
\begin{itemize}
\item Handling systems with large numbers of degrees of freedom. However, this software was not built for this purpose.
\end{itemize}

\item \textbf{Bullet}
\begin{itemize}
\item Comprehensive and accurate documentation.
\item Precision of body interaction
\item Does not apply to my case
\item GPU based simulation pipeline
\item Better explanation of the contact model it uses and extensions to better handle friction between surfaces and preventing all limb detachments and object interpenetrations.
\item Better handling of kinematic chains (open and closed loop).
\item None
\end{itemize}

\item \textbf{crrcsim}
\begin{itemize}
\item - system identification to import real robot models - ROS support - simulated sensors, esp. camera
\end{itemize}

\item \textbf{Drake}
\begin{itemize}
\item We add them as we need them. We link against bullet for collision detection, but otherwise have rolled the entire implementation ourselves.
\end{itemize}

\item \textbf{Dymola}
\begin{itemize}
\item lack of reusable libraries
\end{itemize}

\item \textbf{Gazebo}
\begin{itemize}
\item An up-to-date and complete documentation, up-to-date tutorials.
\item The documentation and possible computation times.
\item physics simulation of non-rigid bodies (Bullet)
\item A better dynamics simulation engine, more realism in general and better look in general of the virtual worlds
\item Easy way (graphic interface) for designing robot model.
\item A model editor.
\item Real-time calculation
\item Say, deformable bodies simulation... (for XDE). AND a decent documentation....
\item a mode to trade off accuracy for speed
\item Software stability, adequate interoperability with ROS.
\item Needs ability to rewind or replay simulations easily.
\item Better time navigation, deformations, cross-platform support
\item Built in model editor
\item Easy and quickly setup of the simulation environment
\end{itemize}

\item \textbf{iCub\_SIM}
\begin{itemize}
\item Possibility to simulate force/torque sensors and skin.
\item Would really like to have a better way to import 3D mesh objects into the iCub simulation. There is some basic functionality, but it only seems to work for very simple shapes.
\end{itemize}

\item \textbf{MapleSim}
\begin{itemize}
\item Good documentation of all possible API commands.
\end{itemize}

\item \textbf{Matlab}
\begin{itemize}
\item easy customization. Currently this is possible, but requieres expert coding habilities.
\end{itemize}

\item \textbf{Moby}
\begin{itemize}
\item Could be more user friendly.
\item It would be great if Moby were better linked with Gazebo or V-Rep.
\end{itemize}

\item \textbf{Morse}
\begin{itemize}
\item Standardization in model representations. (And i do \_not\_ want a "one size fits all" "standard" like URDF! Support for professional communication middleware: HLA, MQP, DDS,... Logging of simulation experiments, in HDF5 files, with a simulation campaign meta model.
\item 2d projection of simulation

\end{itemize}

\item \textbf{MuJoCo}
\begin{itemize}
\item Rendering
\item Inverse dynamics with contacts.
\item None
\end{itemize}

\item \textbf{NeuroRD}
\begin{itemize}
\item Easy way of specifying the model.
\end{itemize}

\item \textbf{ODE}
\begin{itemize}
\item - feedback (that is, accurately simulating force and torque sensors) - confidence estimations of the simulation
\item God's hand: ability to interact with the simulated environment from outside the simulation when we mix simulation and real feedbacks.
\item easy customization. Currently this is possible, but requieres expert coding habilities.
\item easy use of APIs
\item easy customization. Currently this is possible, but requires expert coding abilities.
\item - Supporting robot-model file-formats (e.g. COLLADA, VRML, etc.). - Providing sensor objects (e.g. Gyro, LRF, camera, accelerometer, etc.) - Providing a rendering tool (I want a simple one, such as using OpenGL). - Providing a sophisticated C++ interface.
\item Adding other robots.
\item Easier interfaces and better accuracy.
\item 2. ability to run in a non-gui mode
\end{itemize}

\item \textbf{OpenRave}
\begin{itemize}
\item Good physic engine that is free
\item Simulation of compliant surfaces and contact models.
\item Simulation of compliant surfaces and contact models.
\item Stable physics, more C++ documentation (there are Python examples, which is not bad, but it would be easier the other way around).
\item Better Collision Detection algorithm for non-convex objects, with better force computation. And a more robust and interactive constraint solver.
\item sensors and human
\end{itemize}

\item \textbf{OpenSIM}
\begin{itemize}
\item Environment building. We need something like minecraft for that.
\item An open source gui to create new models. Currently, we have to generate them by writing xml files.
\item lack of rigid contacts
\end{itemize}

\item \textbf{Own code}
\begin{itemize}
\item Flexible part simulation, more realistic contact simulation
\item clean code and documentation.
\end{itemize}

\item \textbf{RCIS}
\begin{itemize}
\item Documentation
\end{itemize}

\item \textbf{roborobo}
\begin{itemize}
\item The fastest the better! And also: (1) parrallel implementation is currently missing (2) save/load snapshot of current state of the simulator (to relaunch from exact same point).
\item - clear documentation
\end{itemize}

\item \textbf{RoboticsLab}
\begin{itemize}
\item None
\item inter-operation with MATLAB
\end{itemize}

\item \textbf{Robotran}
\begin{itemize}
\item 3D collision detection and speed of simulation.
\item A bio-mechanics library
\item Compatibilty towards UNIX.
\end{itemize}

\item \textbf{Simulink + spatial\_v2}
\begin{itemize}
\item geometric models
\end{itemize}

\item \textbf{Simulink and Matlab}
\begin{itemize}
\item being real-time while interacting with real world.
\end{itemize}

\item \textbf{SL}
\begin{itemize}
\item More user-friendly interface.
\item Documentation
\end{itemize}

\item \textbf{trep}
\begin{itemize}
\item We are the developers, so we could add them ourselves.
\end{itemize}

\item \textbf{Vortex}
\begin{itemize}
\item documentation / tutorials
\item Better documentation; more 'end-user' API functionality
\end{itemize}

\item \textbf{VoxCad}
\begin{itemize}
\item Many
\end{itemize}

\item \textbf{V-Rep}
\begin{itemize}
\item none, the forum allows to do feature requests if any, and they usually appear in the next release (once every 2-3 months)
\item Support of more CAD file formats, which are not mesh-based.
\item It would be interesting if it was possible to run the simulator as a web server, so students could run simulations with their controllers over the internet.
\item Lecture material that can be reused to let students use the software in courses. (We are working on this in collaboration with another University though).
\item More friendly interface
\item More general ROS API
\end{itemize}

\item \textbf{Webots}
\begin{itemize}
\item several: easier usage in terms of programming, more simulating robots, more sensor information, better documentation
\item Truly headless operation
\item Non rigid surfaces, fluids
\item having it free and open source would be great of course.
\item 1- being physically realistic. It is ok, but not perfect. 2- access to internal forward dynamics states.
\item fast merge between it and LISP
\item Open source
\end{itemize}

\item \textbf{XDE}
\begin{itemize}
\item Documentation
\item maybe compliant contact, even if soft bodies (cables) exist
\item Documentation
\item Documentation ... Also not sure you can specify an external force as a task in the controller.
\item soft contacts simulation
\end{itemize}

\end{itemize}

\section{Free comments about contact models}

\begin{itemize}

\item \textbf{Autodesk Inventor}
\begin{itemize}
\item You have to determine which contact surface is important for you, then it seems that the software interacts with it similar to a joint.
\end{itemize}

\item \textbf{Morse}
\begin{itemize}
\item Contacts are the last thing I want to see people rely on a simulator! The real world physics is too complex to get into a simulator...
\end{itemize}

\item \textbf{ODE}
\begin{itemize}
\item differential algebraic stiff solvers
\end{itemize}

\item \textbf{OpenRave}
\begin{itemize}
\item ODE/BULLET/PQP flexible implementation.
\end{itemize}

\item \textbf{Robotran}
\begin{itemize}
\item H. Dallali, M. Mosadeghzad, G. Medrano-Cerda, N. Tsagarakis, D. Caldwell, A Dynamic Simulator for the Compliant Humanoid Robot, COMAN, To Appear in ICRA Wokrshop on Developments of Simulation Tools for Robotics \& Biomechanics, Karlsruhe, Germany, May 10, 2013.
\item The contact model I am using has been written by other researchers, and it is not provided with the simulator (robotran).
\end{itemize}

\item \textbf{Simulink + spatial\_v2}
\begin{itemize}
\item physically realistic nonlinear spring+damper in both normal and tangent direction, + clutch in tangent direction implementing genuinely conical friction cone; all changes of state being detected accurately using Simulink's zero-crossing detection.
\end{itemize}

\item \textbf{V-Rep}
\begin{itemize}
\item No
\item I think the most important aspect of contact solving is to get it qualitatively right. At the end of the day, it is only true experimentation in the physical world that will determine if the combined mechanism and controller etc are performing properly.
\end{itemize}

\item \textbf{Webots}
\begin{itemize}
\item I think Webots uses bullets.
\end{itemize}

\item \textbf{XDE}
\begin{itemize}
\item I heard XCD is based on a Gauss-Seidel Algorithm
\end{itemize}

\item \textbf{own software}
\begin{itemize}
\item It is an multi-robot simulator where each robot is simple modelised as an non-honolomic particul.
\end{itemize}
\end{itemize}

\end{document}